%% file: main.tex
\documentclass[sigconf]{acmart}

\AtBeginDocument{%
  }

\setcopyright{acmlicensed}
\copyrightyear{2024} 
\acmYear{2024} 
\setcopyright{acmlicensed}\acmConference[LGM3A '24]{Proceedings of the 2nd Workshop on Large Generative Models Meet Multimodal Applications }{October 28-November 1 2024}{Melbourne, VIC, Australia}
\acmBooktitle{Proceedings of the 2nd Workshop on Large Generative Models Meet Multimodal Applications (LGM3A '24), October 28-November 1 2024, Melbourne, VIC, Australia}
\acmDOI{10.1145/3688866.3689124}
\acmISBN{979-8-4007-1193-0/24/10}

\input{preamble}
\begin{document}

\title[Geo-LLaVA: A Large Multi-Modal Model for Solving Geometry Math Problems \\ with Meta In-Context Learning]{Geo-LLaVA: A Large Multi-Modal Model for Solving Geometry Math Problems with Meta In-Context Learning}
\author{Shihao Xu}
\authornote{Both authors contributed equally to this research.}
\email{xush0019@ntu.edu.sg}
\orcid{0000-0001-8295-5692}

\affiliation{%
  \institution{Huawei Singapore Research Center}
  \city{Singapore}
  \country{Singapore}
}

\author{Yiyang Luo}
\authornotemark[1]
\email{luoyiyang2@huawei.com}
\orcid{0009-0008-8094-180X}

\affiliation{%
  \institution{Huawei Singapore Research Center}
  \city{Singapore}
  \country{Singapore}
}

\author{Wei Shi}
\email{w.shi@huawei.com}

\affiliation{%
  \institution{Huawei Singapore Research Center}
  \city{Singapore}
  \country{Singapore}
}

\renewcommand{\shortauthors}{Shihao Xu, Yiyang Luo, \& Wei Shi}

\input{sec/0_abstract}




\maketitle

\input{sec/1_intro}

\input{sec/2_related}
\input{sec/3_method}
\input{sec/4_experiment}
\input{sec/6_conclution}

\bibliographystyle{acm}
\balance
\bibliography{main}

\end{document}

%% file: preamble.tex
\usepackage{xspace}
\usepackage{diagbox}
\usepackage{bm}
\usepackage{multirow}
\usepackage{mdframed}
\usepackage[inline]{enumitem}
\usepackage{caption}
\usepackage{subcaption}

\makeatletter
\DeclareRobustCommand
  \myvdots{\vbox{\baselineskip2\p@ \lineskiplimit\z@
    \hbox{.}\hbox{.}\hbox{.}}}
\makeatother

\makeatletter

%% file: sec/0_abstract.tex
\begin{abstract}
Geometry mathematics problems pose significant challenges for large language models (LLMs) because they involve visual elements and spatial reasoning. 
Current methods primarily rely on symbolic character awareness to address these problems. 
Considering geometry problem solving is a relatively nascent field with limited suitable datasets and currently almost no work on solid geometry problem solving, we collect a geometry question-answer dataset by sourcing geometric data from Chinese high school education websites, referred to as GeoMath. It contains solid geometry questions and answers with accurate reasoning steps as compensation for existing plane geometry datasets.
Additionally, we propose a Large Multi-modal Model (LMM) framework named Geo-LLaVA, which incorporates retrieval augmentation with supervised fine-tuning (SFT) in the training stage, called meta-training, and employs in-context learning (ICL) during inference to improve performance.
Our fine-tuned model with ICL attains the state-of-the-art performance of 65.25\% and 42.36\% on selected questions of the GeoQA dataset and GeoMath dataset respectively with proper inference steps. Notably, our model initially endows the ability to solve solid geometry problems and supports the generation of reasonable solid geometry picture descriptions and problem-solving steps. Our research sets the stage for further exploration of LLMs in multi-modal math problem-solving, particularly in geometry math problems.
\end{abstract}

%% file: sec/1_intro.tex
\section{Introduction}
\label{sec:intro}

\begin{figure}[t]
    \centering
    \includegraphics[width=\linewidth]{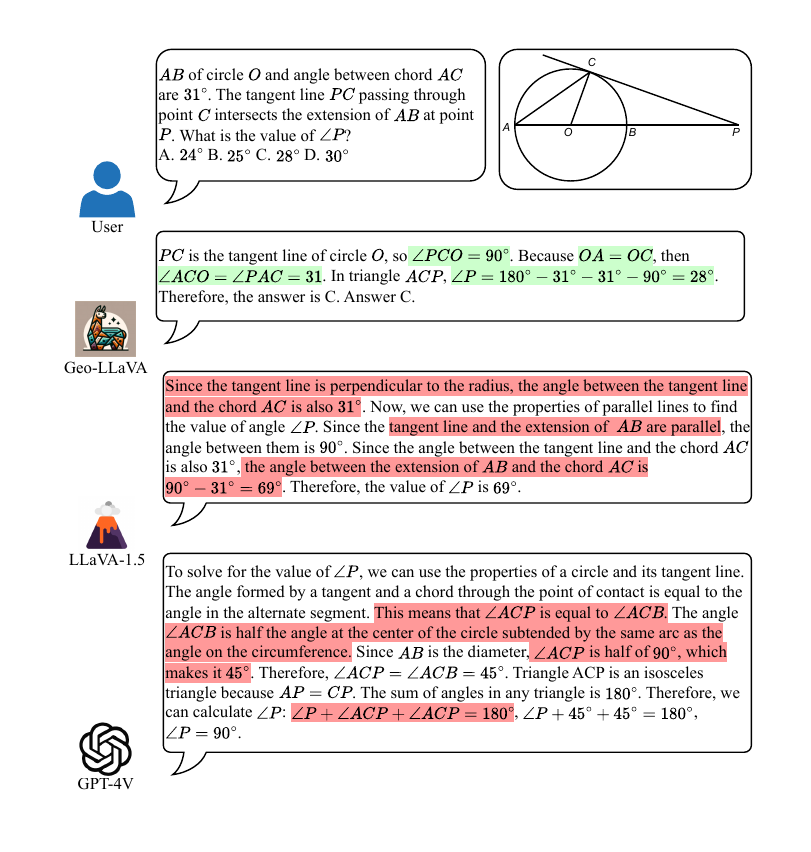}
    \caption{Example of geometry question answering. LLaVA-1.5 often provides a vague image description, omitting many details and leading to wrong answers. GPT-4V struggles to understand geometry graphs, resulting in incorrect answers. In contrast, our Geo-LLaVA offers concise and accurate solutions to geometry problems.}
    \Description{Image that illustrates the samples for solving geometry questions by LLMs.}
    \label{fig:examples}
\end{figure}

There has been an increased interest in using deep learning models, especially LMMs, for addressing computer vision challenges recently. Aligning image adapters with large language models has achieved remarkable success in image captioning and visual question answering (VQA), highlighting their powerful reasoning and visual understanding capabilities \cite{bordes_introduction_2024}.
Despite their accomplishments in these areas, there has been limited investigation into utilizing LLMs for more complex multi-modal math problems, particularly geometry-related ones. 


The debate between symbolic and probabilistic approaches in mathematical reasoning persists. Traditionally, solving geometry problems involves analyzing diagrams and texts, converting them into logical expressions using formal symbolic language, and applying predefined geometry theorems to find solutions \cite{chen2021geoqa, zhang2023multi}. Alternatively, geometry problem-solving can be viewed as text generation with multi-modal input, which is more generalized and applicable to a broader range of mathematical problems, including trigonometry and vector graphics. By examining probabilistic approaches, we can better understand their strengths and limitations, leading to more effective strategies for solving geometry problems.

Current Large Multi-modal Models (LMMs) have shown promising capabilities in visual understanding and question-answering tasks, as demonstrated by models such as BLIP-2 \cite{li2023blip}, LLaVA \cite{llava}, Flamingo \cite{alayrac2022flamingo}, MiniGPT4 \cite{zhu2023minigpt}, and InstructBLIP \cite{instructblip}. However, these models still lack a deep comprehension of geometry images, which is crucial for solving geometry problems. Additionally, existing small language models often lack the mathematical reasoning abilities to solve complex math problems effectively. 

In this paper, we propose a solution to the challenges of multi-modal math problem-solving by introducing an LLM framework called Geo-LLaVA. 
The main contributions of this paper include: 
\begin{itemize}
    \item We form a geometry question-answer dataset, GeoMath, with reasoning steps from Chinese high school education websites and expand the dataset by collecting more data from existing datasets such as GeoQA+ \cite{chen2021geoqa}, Unigeo \cite{chen2022unigeo}, and PGPS9K \cite{zhang2023multi}, and creating reasoning steps for them.
    
    \item We employ a new LLM framework named Geo-LLaVA with around 13 billion parameters. It can effectively generate reasoning steps and answers, which is the first model exploring both plane and solid geometry problems. 
    
    \item As far as we know, this is the first multimodal meta-training \cite{min2021metaicl} method to enhance geometry problem-solving ability, which shows the state-of-the-art performance on multiple geometry datasets, including GeoQA+ and GeoMath.
    
\end{itemize}

%% file: sec/2_related.tex
\section{Related Work}
\label{sec:related_work}

\paragraph{Meta-Training Approaches}

The general issue of meta-training \cite{schmidhuber1987evolutionary}, which encompasses few-shot learning, has been studied for numerous years. Recently, meta-training has emerged as a significant technique in machine learning. Using prior knowledge, it aims to create models that can swiftly adapt to new tasks with minimal data. meta-training is particularly pertinent in geometry problem-solving, where the diversity of problems can vary greatly.
The most recent meta-training models, such as MAML \cite{finn2017model}, Hypernetworks \cite{ha2022hypernetworks}, and \cite{ravi2016optimization}, have shown promise in rapidly adapting to new tasks with few-shot learning capabilities.

\paragraph{Multi-Modal Large Language Model}
Concurrently, the success of LLMs has inspired investigations into vision-language interaction, resulting in the development of multi-modal large language models (MLLMs) \cite{llava, li2023blip, instructblip, weng2022large, 2023GPT4VisionSC, bai2023qwen}. These models have demonstrated remarkable abilities in generating detailed descriptions and engaging in dialogue based on visual inputs. Nonetheless, we observe that even the most advanced MLLMs struggle with resolving geometric problems using diagrams and figures.

\paragraph{Geometry Problem Solving by LLMs}
Recently, math-specific LLMs such as Llemma \cite{azerbayev2023llemma} and Mathcoder \cite{wang2023mathcoder} have shown significant capabilities in text-only mathematical reasoning tasks and are competitive with general large language models like GPT-4 \cite{achiam2023gpt} and PaLM-2 \cite{anil2023palm} on a much smaller scale. Notably, AlphaGeometry \cite{Trinh2024AlphaGeo} has exhibited impressive performance in solving challenging geometry problems, though it cannot process images and must rely on text descriptions. Current math-specific multimodal models, such as G-llava \cite{gao2023g}, UniMath \cite{liang2023unimath}, and UniChart \cite{masry2023unichart}, are primarily focused on plane geometry or chart-based problems and still lag behind general multimodal models such as GPT-4V \cite{2023GPT4VisionSC} in benchmark testing. Yet, no works have utilized RAG or meta-training techniques on geometric problem-solving.

%% file: sec/3_method.tex
\section{Method}

The Geo-LLaVA model, illustrated in Figure \ref{fig:examples}, consists of a retrieval network and an LMM backbone. The retrieval network's role is to fetch similar questions and their solutions as in-context samples during the training and inference phases. In the following section, we offer a comprehensive overview of the design process for our retrieval network. Next, we delve into our method of fine-tuning the model using image captioning, question-answering, and geometry math-solving datasets through Meta in-context learning. Lastly, we will describe how we have integrated a multi-modal chain of thoughts (CoT) during the inference phase to boost the model's performance even further.

\begin{figure*}
  \centering
  \includegraphics[width=0.9\linewidth]{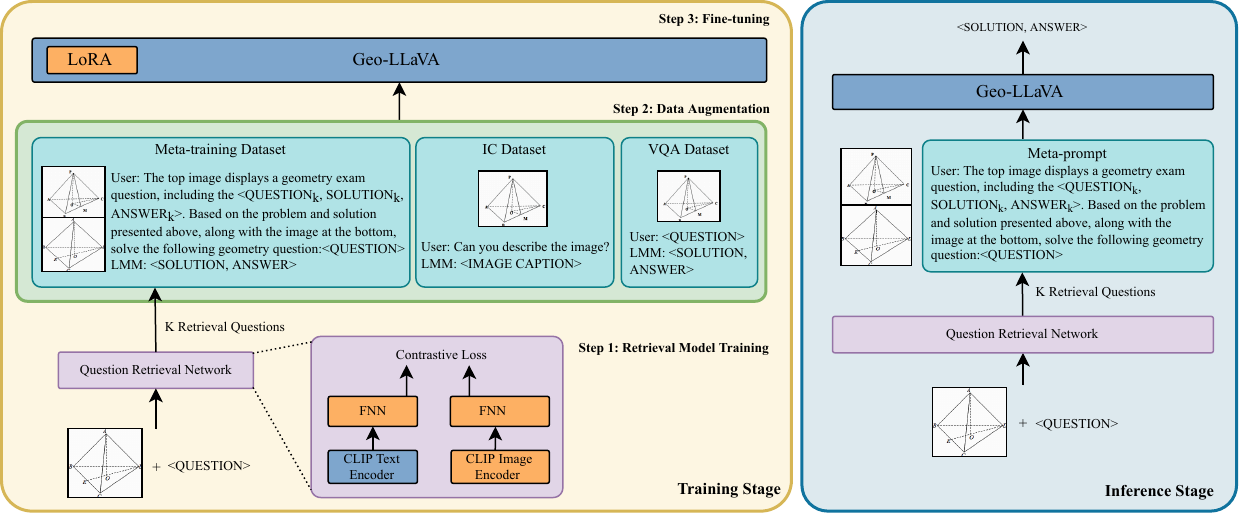}
  \caption{The overview of Geo-LLaVA model architecture. The training process includes three steps: training the retrieval model, augmenting data with the meta-training dataset, and fine-tuning the Language Model with the LoRA module. The question retrieval network is trained using inputs from a CLIP text encoder and a CLIP image encoder, supervised by a contrastive loss. FNNs represent feedforward networks. During the fine-tuning step, a LoRA \cite{hu2021lora} is introduced for efficient fine-tuning and improved results.}
  \label{fig:overview}
  \Description{Image that illustrates the overall structure of the Geo-LLaVA.}
\end{figure*}

\subsection{Retrieval Network}

In this study, similar to CLIP \cite{clip2021}, we implemented a dual-tower network framework for retrieval tasks. Specifically, the pre-trained ViT-L-14 \cite{dosovitskiy2020image} and Bert \cite{devlin2018bert} (Bert-base-uncased) models were applied as the image and language encoders, respectively. We integrated two adapter layers into each encoder to ensure compatibility between these encoders. These adapter layers comprise three linear layers with ReLU activation functions, designed to harmonize the embedding dimensions of both encoders.

The Bert model \cite{devlin2018bert} was chosen as the language encoder due to its strong performance in natural language processing tasks. Importantly, we transformed math-specific tokens absent in the BERT pre-trained vocabulary into words (e.g., $\triangle \rightarrow$ triangle, $\perp \rightarrow$ perpendicular to) during the preprocessing stage. We believe the BERT model can effectively grasp the meaning of inputs related to geometric problems. 
Similarly, the ViT-L-14 \cite{dosovitskiy2020image} model was selected as the image encoder for its outstanding performance in image recognition tasks. This model employs a transformer-based architecture, which is particularly adept at processing visual information. However, the pre-trained ViT model may not generalize well to geometric math images. As a result, we retrained the parameters of the two adapter layers and the ViT model from scratch using question-image pairs with the InfoNCE loss, a contrastive learning technique recognized for its efficacy in training neural networks for retrieval tasks. This loss function encourages the model to learn meaningful representations of the input question and the corresponding image, promoting accurate information retrieval.

\subsection{LMM backbone}

This study utilized the pre-trained LLaVA1.5-13B \cite{llava} as the LMM backbone. 
This model leverages the renowned LLaMA-2 \cite{touvron2023llama2} for advanced language processing tasks and incorporates the CLIP \cite{clip} visual encoder ViT-L/14 \cite{dosovitskiy2020image} for sophisticated visual comprehension. The integration involves a Multi-Layer Perceptron (MLP) based vision-language connector that aligns the outputs of the vision encoder with the language model. This alignment is crucial as it significantly enhances the model to handle and understand multi-modal data effectively.

\subsection{Datasets for Multi-Modal Geometric Concepts and Reasoning}
To address the limitations of existing models in understanding and reasoning about multi-modal geometric concepts, we developed three specialized geometry datasets:
\begin{itemize}
\item GeoMath-IC (Image-Context): This dataset includes images paired with simple yet comprehensive descriptions and aims to bridge the gap between visual inputs and textual descriptions in geometric concepts.
\item GeoMath-QA (Question-Answer): This dataset focuses on providing questions with detailed reasoning steps. The questions are designed to challenge the model's understanding and reasoning capabilities regarding geometric concepts.
\item GeoMath-Meta: This dataset includes a range of geometry problems that require the model to generalize from its learning on GeoMath-IC and GeoMath-QA.
\end{itemize}
These datasets were subsequently used to fine-tune the model using Low-Rank Adaptation (LoRA) \cite{hu2021lora}.  
Detailed information about the dataset creation process and examples can be found in Section \ref{sec:data_augmentation}.

We adopted the input format from the original Flamingo model, which efficiently integrates visual and linguistic elements to enhance their synergistic interaction.
This structured input format involves specific templates and prompts that guide the model in understanding the context and relationships between the visual and textual components.

\subsection{Enhancing In-Context Learning}

To further improve the model's in-context learning capabilities, especially for smaller models, we explored using meta-training \cite{min2021metaicl}. This approach enhances the performance of the model by providing relevant context during the fine-tuning stage. The procedure involves the following steps: 1) Contextual Retrieval: For each input sample, we retrieved the $K$ most similar samples from the training data, ensuring that the input sample itself is excluded, where $K$ is set to $1$ in this paper. This retrieval is based on semantic similarity metrics, ensuring the context is highly relevant. 2) Concatenation and Fine-Tuning: The retrieved texts and corresponding images are concatenated with the input sample. This concatenated form is used to fine-tune the LMM, aligning it better with the reasoning required for tasks. Since the pre-trained LLaVA only supports single-image input, we vertically merge $K$ images into a single image.

The detailed methodologies and the careful orchestration of these components can be referenced throughout the underlying sections of the study, particularly in Section \ref{sec:data_augmentation} and Figure \ref{fig:overview}.

% ...


%% file: sec/4_experiment.tex
\section{Experiment}

\subsection{Dataset}
\label{sec:Dataset}

In this study, we collected about 10K solid geometry multimodal QA datasets from the 21st-century education website in China, named GeoMath dataset. All Chinese contents in these datasets are automatically translated to English using ChatGPT3.5. The detailed statistics of these three datasets are shown in Table 1. In addition, to supplement the data of solid geometry, we formed the geometry data from two existing datasets (GeoQA+ \cite{cao2022augmented} and PSDK9K), which include images of plane geometry as well as questions and answers.

\begin{table}[htbp]
  \centering
  \caption{The detailed statistics of four datasets. Each row represents the number of samples of specific types. The number inside the brackets represents the number of test samples.}
  \scalebox{0.87}{
    \begin{tabular}{c|cccc}
    \toprule
    \diagbox{Stat. type}{Dataset} & GeoQA & PSDK-9K & UniGeo & GeoMath \\
    \midrule
    QA-selection & 12526(1509) & 9986(1047) & - & 4258(404) \\
    QA-cloze & - & - & - & 1423 (150) \\
    QA-proving & - & - & 9309(2899) & 3474 (352) \\
    \midrule
    Image-text pairs & 4406 & 4000 & - & 4540 (453) \\
    Provide solution & $\surd$ & - & $\surd$ & $\surd$ \\
    \bottomrule
    \end{tabular}}
  \label{tab:datasets}
\end{table}

\subsection{Setup Details}

\subsubsection{Data Augmentation}
\label{sec:data_augmentation}

Paraphrasing by LLMs can generate a more diverse set of training examples and has been widely used for data augmentation. 
Similarly, we adopt GPT3.5 \cite{achiam2023gpt} for image caption and question-answering samples followed by a translation from Chinese to English and employ text rewriting to increase variety. 
We utilize an LLM to rephrase the input text in 5 different ways, resulting in a six times larger sample size for image caption and QA. For the GeoMath-Meta dataset, the retrieval model selects the top 5 samples with the highest similarity to construct the data for Metatraining. This ensures consistency in the quantity of the final samples and QA pairs.

\subsubsection{Training settings}
We select LLaVA-1.5 \cite{llava}, a large multimodal model that combines the strengths of LLaMA-2 \cite{touvron2023llama2} and fine-tuned retrieval model
\cite{clip}, for our experiments. All experiments are conducted with consistent parameter settings during the LoRA fine-tuning phase. Specifically, we use a learning rate of $2 \times 10^{-4}$ with a cosine learning rate scheduler and train the model for 5 epochs. The maximum token generation length is 2048. The batch size per GPU is 4, and we use gradient accumulation steps set to 4. We initially evaluate the experiments on the validation set to identify the best results, which are subsequently tested on the test set.



\begin{table}[]
\caption{
The accuracy of various models across different datasets. Additionally, an ablation study is presented, illustrating the sequential application of different techniques for solving geometry mathematics problems. IC, QA, and MT stand for adding the GeoMath-IC, GeoMath-QA, and GeoMath-Meta datasets for training respectively. We indicate whether providing few-shot samples in the inference phase using 'with ICL' and 'w/o ICL'.}

\scalebox{0.8}{
\begin{tabular}{l|cc}
\toprule
 Models & GeoQA+ (\%,$\uparrow$) & GeoMath (\%,$\uparrow$) \\
\midrule
 Openflamingo-9B \cite{awadalla2023openflamingo} & 27.37 & 21.14 \\
 LLaVA-1.5-13B \cite{llava} & 29.30 & 22.28\\
 Bard \cite{google_bard}  & 47.10 & 20.00 \\
 GPT-4V \cite{2023GPT4VisionSC} & 50.50 & 20.00 \\
 DPE-NGS \cite{cao2022augmented} & 66.09 & -  \\
 G-llava-13B \cite{gao2023g} & \textbf{67.00} & - \\
\midrule
Geo-LLaVA-13B (QA) & 57.70 & 28.60 \\
Geo-LLaVA-13B (IC+QA) & 61.04 (+3.34) & 37.12 (+8.52) \\
Geo-LLaVA-13B (IC+QA+MT, w/o ICL) & 63.13 (+2.09) & 41.48 (+4.36) \\
Geo-LLaVA-13B (IC+QA+MT, with ICL) & 65.25 (+2.12) & \textbf{42.36} (+0.88) \\
\bottomrule
\end{tabular}}
\label{tab:main_exp}
\end{table}

\subsection{Experiment results}
Table \ref{tab:main_exp} summarizes the main results on GeoQA+ and GeoMath datasets. Three small-size LMMs (G-llava \cite{gao2023g}, OpenFlamingo \cite{awadalla2023openflamingo} and LLaVA \cite{llava}) for both zero-shot and finetuning and two extremely large LMMs (Bard \cite{google_bard} and ChatGPT-4V \cite{2023GPT4VisionSC}) are chosen as baseline models. We finetuned the model 5 times to calculate the mean and standard deviation of the evaluation metric. 


The experiment results, summarized in Table \ref{tab:main_exp}, demonstrate the performance of various models across the GeoQA+ and GeoMath datasets, alongside an ablation study exploring the incremental application of techniques in our model. The proposed Geo-LLaVA-13B model showed significant improvements compared to the GPT-4V and Bard through the sequential addition of the GeoMath-IC, GeoMath-QA, and GeoMath-Meta datasets, with final accuracies of 65.25\% and 42.36\% on the respective datasets.

The results indicate that the proposed Geo-LLaVA-13B model outperforms several other models in solving geometry mathematics problems, particularly when using a combination of the GeoMath-IC, GeoMath-QA, and GeoMath-Meta datasets on the solid geometric problems. The step-by-step improvements highlight the effectiveness of sequentially incorporating these datasets and techniques. Additionally, the model's performance further benefits from ICL, suggesting that providing few-shot examples during inference significantly enhances its accuracy. 

%% file: sec/6_conclution.tex
\section{Conclusion}
In conclusion, this paper presents a novel approach to address the challenges inherent in multi-modal math problem-solving on geometry. Our research emphasizes the pivotal role of integrating meta-learning into models, enabling them to accurately interpret and reason through complex visual and textual inputs—an essential capability for resolving geometric problems effectively. Recognizing the limitations posed by the small size and the absence of reasoning steps in current geometry datasets, we have developed a pioneering dataset called GeoMath. This dataset amalgamates reasoning steps drawn from pre-existing datasets and materials sourced from a Chinese high school educational website, thereby filling a critical gap in available resources.

This study not only provides significant contributions to the domain of multi-modal geometry problem-solving but also paves the way for future research endeavors. The application of LLMs and LMMs to this domain promises to unlock new potential and methodologies.
The development of GeoMath stands as a notable milestone, offering robust strategies for addressing the complexities of geometry problems.
Future research could explore refining these models further, expanding the dataset with additional problem types, and investigating their applicability across various educational contexts, potentially transforming how multi-modal mathematical reasoning is approached in both academic and practical settings.